\pdfoutput=1

\documentclass[11pt, dvipsnames]{article}

\usepackage[]{EMNLP2022}

\usepackage{times}
\usepackage{latexsym}

\usepackage[T1]{fontenc}

\usepackage[utf8]{inputenc}

\usepackage{microtype}

\usepackage{enumitem}
\setitemize{noitemsep,topsep=0pt,parsep=0pt,partopsep=0pt}

\usepackage{inconsolata}
\usepackage{booktabs}
\usepackage{multirow}
\usepackage{amsmath}
\usepackage{amsfonts}
\usepackage{amsthm}
\usepackage{pifont}
\usepackage{xspace}
\usepackage{graphicx}
\usepackage{bm}
\usepackage{color, colortbl}
\usepackage{csquotes}
\usepackage{anyfontsize}
\usepackage{comment}
\usepackage{float}

\usepackage{titlesec}
\titlespacing*{\section}{0pt}{1.6ex plus 0.6ex minus 0.3ex}{1.3ex plus 0.5ex minus 0.2ex}
\titlespacing*{\subsection}{0pt}{0.8ex plus 0.5ex minus 0.2ex}{0.5ex plus 0.5ex minus 0.2ex}

\def\middlecol{\hskip 6pt}
\def\smallcol{\hskip 4pt}
\def\tinycol{\hskip 2pt}

\newcommand{\ours}{ReAtt\xspace}

\newcommand{\adjust}{adjust\xspace}
\newcommand{\adjustment}{cross-document adjustment\xspace}
\newcommand{\adjustloss}{cross-document adjustment loss\xspace}

\newcommand{\cmark}{\ding{51}}
\newcommand{\xmark}{\ding{55}}

\newcommand{\topk}{\text{topk}}
\newcommand{\softmax}{\text{softmax}}
\newcommand{\kl}[2]{\text{KL}\Big(#1\Big|\Big|#2\Big)}

\newcommand{\msm}{MS MARCO\xspace}
\newcommand{\improve}[1]{$_{\text{+#1}}$}
\newcommand{\squad}{\hspace{0.5em}} 

\definecolor{LightCyan}{rgb}{0.88,1,1}
\newcommand{\hlcell}{\cellcolor{LightCyan!50}}
\newcommand{\hlrow}{\rowcolor{LightCyan!50}}
\newcommand{\dlrow}{\rowcolor{gray!20}}

\newcommand{\better}[1]{\textcolor{Green}{#1}}
\newcommand{\worse}[1]{\textcolor{Maroon}{#1}}

\newcommand{\supervised}[1]{\textcolor{Maroon}{#1}}
\newcommand{\pretrain}[1]{\textcolor{Green}{#1}}
\newcommand{\adaptation}[1]{\textcolor{Orange}{#1}}

\title{Retrieval as Attention: End-to-end Learning of Retrieval \\ and Reading within a Single Transformer}

\author{Zhengbao Jiang$^{\heartsuit}$\thanks{~~The first two authors contributed equally.}, \quad Luyu Gao$^\heartsuit$\footnotemark[1], \quad Jun Araki$^\diamondsuit$, \quad Haibo Ding$^\diamondsuit$\thanks{~~Haibo Ding is now at Amazon.}, \\
{ \bf Zhiruo Wang$^\heartsuit$, \quad Jamie Callan$^\heartsuit$, \quad Graham Neubig$^\heartsuit$ } \\
$^\heartsuit$Language Technologies Institute, Carnegie Mellon University \\
$^\diamondsuit$Bosch Research North America \\
\texttt{\{zhengbaj,luyug,zhiruow,callan,gneubig\}@cs.cmu.edu} \\
\texttt{\{jun.araki,haibo.ding\}@us.bosch.com}}

\begin{document}
\maketitle
\begin{abstract}
Systems for knowledge-intensive tasks such as open-domain question answering (QA) usually consist of two stages: efficient retrieval of relevant documents from a large corpus and detailed reading of the selected documents to generate answers.
Retrievers and readers are usually modeled separately, which necessitates a cumbersome implementation and is hard to train and adapt in an end-to-end fashion.
In this paper, we revisit this design and eschew the separate architecture and training in favor of a single Transformer that performs \textbf{Re}trieval as \textbf{Att}ention (\textbf{\ours}), and end-to-end training solely based on supervision from the end QA task.
We demonstrate for the first time that a single model trained end-to-end can achieve both competitive retrieval and QA performance, matching or slightly outperforming state-of-the-art separately trained retrievers and readers.
Moreover, end-to-end adaptation significantly boosts its performance on out-of-domain datasets in both supervised and unsupervised settings, making our model a simple and adaptable solution for knowledge-intensive tasks.
Code and models are available at \url{https://github.com/jzbjyb/ReAtt}.
\end{abstract}

\section{Introduction}
Knowledge-intensive tasks such as question answering (QA), fact checking, and dialogue generation require models to gather relevant information from potentially enormous knowledge corpora (e.g., Wikipedia) and generate answers based on gathered evidence.
A widely used solution is to first \emph{retrieve} a small number of relevant documents from the corpus with a \emph{bi-encoder} architecture which encodes queries and documents independently for efficiency purposes, then \emph{read} the retrieved documents in a more careful and expansive way with a \emph{cross-encoder} architecture which encodes queries and documents jointly \cite{orqa-2019-lee,realm-2020-guu,rag-2020-lewis,atlas-2022-izacard}.
The distinction between retrieval and reading leads to the widely adopted paradigm of treating retrievers and readers separately.
Retrievers and readers are usually two separate models with heterogeneous architectures and different training recipes, which is cumbersome to train.
Even though two models can be combined in an ad-hoc way for downstream tasks, it hinders effective end-to-end learning and adaptation to new domains.

There have been several attempts to connect up reader and retriever training \cite{orqa-2019-lee,realm-2020-guu,rag-2020-lewis,emdr2-2021-sachan,yono-2021-lee,atlas-2022-izacard}.
However, retrievers in these works are not learned in a fully end-to-end way.
They require either initialization from existing supervisedly trained dense retrievers \cite{rag-2020-lewis}, or expensive unsupervised retrieval pretraining as warm-up \cite{orqa-2019-lee,realm-2020-guu,emdr2-2021-sachan,yono-2021-lee,atlas-2022-izacard}.
The reliance on retrieval-specific warm-up and the ad-hoc combination of retrievers and readers makes them less of a unified solution and potentially hinders their domain adaptation ability.
With the ultimate goal of facilitating downstream tasks, retriever and reader should instead be fused more organically and learned in a fully end-to-end way.

In this paper, we focus on one of the most important knowledge-intensive tasks, open-domain QA.
We ask the following question: is it possible to perform both retrieval and reading \emph{within a single Transformer model}, and train the model in a \emph{fully end-to-end} fashion to achieve competitive performance from both perspectives?
Such a single-model end-to-end solution eliminates the need for retrieval-specific annotation and warm-up and simplifies retrieval-augmented training, making adaptation to new domains easier.
Based on the analogy between self-attention which relates different tokens in a single sequence \cite{transformer-2017-vaswani} and the goal of retrieval which is to relate queries with relevant documents, we hypothesize that self-attention could be a natural fit for retrieval, and it allows an organic fusion of retriever and reader within a single Transformer.

\begin{figure}[tb]
\includegraphics[width=1.0\columnwidth, clip, keepaspectratio]{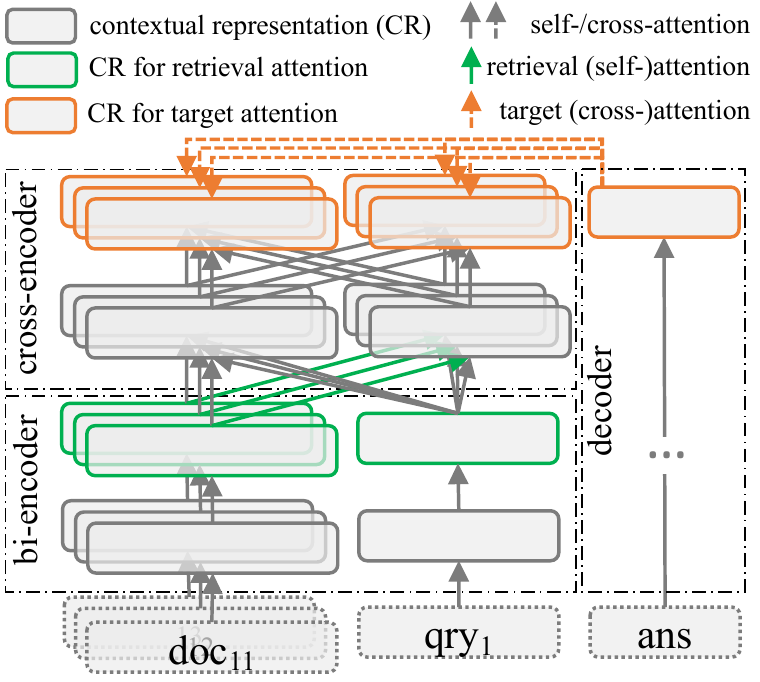}
\centering
\caption{Illustration of Retrieval as Attention (\ours) with the first $B$=2 encoder layers as bi-encoder (i.e., retriever) and the rest $L$-$B$=2 layers as cross-encoder. During training, the \textcolor{Green}{retrieval attention} between a query $\bm{q}_1$ and documents $\bm{d}_{11,12,13}$ is adjusted by minimizing its discrepancy from the \textcolor{Orange}{target attention}.
For simplicity, we use a single arrow to represent attention of a single head between multiple tokens.}
\label{fig:reatt}
\end{figure}

Specifically, we start from a encode-decoder T5 \cite{t5-2020-raffel} and use it as both retriever and reader.
We use the first $B$ encoder layers as bi-encoder to encode queries and documents independently, and the attention score at layer $B+1$ (denoted as \emph{retrieval attention}) to compute relevance scores, as shown in \autoref{fig:reatt}.
We found that directly using self-attention for retrieval underperforms strong retrievers, which we conjecture is because self-attention pretrained on local context is not sufficient to identify relevant information in the large representation space of the whole corpus.
To solve this, we propose to compute retrieval attention between a query and a large number of documents and \emph{\adjust the retrieval attention across documents}.
For each query, we compute retrieval attention over both close documents that potentially contain positive and hard negative documents, and documents of other queries in the same batch as random negatives.
The retrieval attention is adjusted by minimizing its discrepancy from the cross-attention between the decoder and encoder (denoted as \emph{target attention}), which is indicative of the usefulness of each document in generating answers \cite{fid-kd-2021-izacard}.
The resulting \textbf{Re}trieval as \textbf{Att}ention model (\textbf{\ours}) is a single T5 trained based on only QA annotations and simultaneously learns to promote useful documents through \adjustment.

We train \ours on Natural Questions dataset (NQ) \cite{nq-2019-kwiatkowski} in a fully end-to-end manner.
It achieves both competitive retrieval and QA performance, matching or slightly outperforming state-of-the-art retriever ColBERT-NQ \cite{colbertnq-2020-khattab} trained with explicit retrieval annotations and strong QA model FiD \cite{fid-2021-izacard,fid-kd-2021-izacard}, demonstrating for the first time end-to-end training can produce competitive retriever and reader within a single model.
To further test \ours's generalization and end-to-end adaptation ability, we conduct zero-shot, supervised, and unsupervised adaptation experiments on 7 datasets from the BEIR benchmark \cite{beir-2021-thakur}.
In all settings, end-to-end adaptation improves the retrieval performance usually by a large margin, achieving comparable or superior performance to strong retrieval adaptation and pretraining methods.

\section{Retrieval as Attention (\ours)}\label{sec:reatt}
With the goal of developing a single Transformer that can perform both retrieval and reading, and the analogy between retrieval and self-attention, we first introduce architecture changes to allow retrieval as attention (\autoref{sec:reatt_intro}), then examine how well attention as-is can be directly used to perform retrieval (\autoref{sec:reatt_asis}).

\subsection{Formal Definition}\label{sec:reatt_definition}
We first briefly define the task of retrieval and question answering.
As mentioned in the introduction, queries and documents need to be represented independently for efficient retrieval which implies a bi-encoder architecture that has no interaction between queries and documents.
Without loss of generality, we use $E_{\bm{d}}=\text{biencoder}(\bm{d})$ to denote one or multiple representations generated by a bi-encoder based on a document from a corpus $\bm{d} \in \mathcal{D}$, and likewise $E_{\bm{q}}=\text{biencoder}(\bm{q})$ to denote query representations.\footnote{Queries and documents can use different bi-encoders but we use one notation for simplicity.}
The top-k documents most relevant to a query are retrieved by $\mathcal{D}^{\text{ret}}_{\bm{q}} = \arg\topk_{\bm{d} \in \mathcal{D}}r(E_{\bm{q}}, E_{\bm{d}})$, where function $r$ computes relevance based on query and document representations which can be as simple as a dot product if queries and documents are encoded into a single vector, and $\mathcal{D}^{\text{ret}}_{\bm{q}}$ stands for the returned documents.
We consider encoder-decoder-based generative question answering in this paper, which jointly represents queries and retrieved documents with the encoder $E_{\bm{q},\bm{d}}=\text{crossencoder}(\bm{q}, \bm{d})$, and generates the answer $\bm{a}$ autoregressively with the decoder $P^{\text{gen}}(\bm{a}|\bm{q},\bm{d})=P^{\text{gen}}(\bm{a}|E_{\bm{q},\bm{d}})$.
To handle multiple retrieved documents, we follow the fusion-in-decoder model (FiD) \cite{fid-2021-izacard} which encodes each query-document pair independently and fuse these representations in decoder through cross-attention $P^{\text{gen}}(\bm{a}|\bm{q},\mathcal{D}^{\text{ret}}_{\bm{q}})=P^{\text{gen}}(\bm{a}|E_{\bm{q},\bm{d}_1},...,E_{\bm{q},\bm{d}_{|\mathcal{D}^{\text{ret}}_{\bm{q}}|}})$.
Negative log likelihood (NLL) is used in optimization $\mathcal{L}_{\text{QA}}=-\log P^{\text{gen}}(\bm{a}|\bm{q},\mathcal{D}^{\text{ret}}_{\bm{q}})$.

\subsection{Leveraging Attention for Retrieval}\label{sec:reatt_intro}
Next, we introduce our method that directly uses self-attention between queries and documents as retrieval scores.

\paragraph{Putting the Retriever into Transformers}
As illustrated in \autoref{fig:reatt}, we choose T5 \cite{t5-2020-raffel} as our base model, use the first $B$ layers of the encoder as the \emph{bi-encoder ``retriever''} by disabling self-attention between queries and documents, and the remaining $L-B$ layers as the \emph{cross-encoder ``reader''}.
We use the self-attention paid from query tokens to document tokens at the $B+1$-th layer as the retrieval score, which is denoted as \emph{retrieval attention} (green arrows in \autoref{fig:reatt}).
It is computed based on the independent query and document contextual representations from the last ($B$-th) layer of the bi-encoder (green blocks in \autoref{fig:reatt}).
Formally for an $H$-head Transformer, document and query representations are:
\begin{align*}
E_{\bm{d}} &= \{ K^{B+1,h}_{\bm{d}} \in \mathbb{R}^{|\bm{d}| \times e} \}_{h=1}^{H}, \\
E_{\bm{q}} &= \{ Q^{B+1,h}_{\bm{q}} \in \mathbb{R}^{|\bm{q}| \times e} \}_{h=1}^{H}, 
\end{align*}
where $K$ and $Q$ are key and query vectors of the token sequence used in self-attention, $|\bm{d}|$ and $|\bm{q}|$ are document and query length, and $e$ is the dimensionality of each head.
The retrieval attention matrix from query tokens to document before softmax for one head is computed by:
\begin{equation*}
A^{B+1,h}_{\bm{q},\bm{d}} = Q^{B+1,h}_{\bm{q}} \times {K^{B+1,h}_{\bm{d}}}^T \in \mathbb{R}^{|\bm{q}| \times |\bm{d}|}.
\end{equation*}
Directly using attention for retrieval can not only leverage its ability to identify relatedness, it is also a natural and simple way to achieve both retrieval and reading in a single Transformer with minimal architectural changes, which facilitates our final goal of end-to-end learning.

\paragraph{From Token Attention to Document Relevance}
Given the token-level attention scores $A^{B+1,h}_{\bm{q},\bm{d}}$, the relevance between $\bm{q}$ and $\bm{d}$ is computed by avg-max aggregation: choosing the most relevant document token for each query token (i.e., max) then averaging across query tokens:
\begin{equation}\label{eq:avgmax}
r_h(\bm{q}, \bm{d}) = \text{avg}_{0}\big(\text{max}_1(A^{B+1,h}_{\bm{q},\bm{d}})\big),
\end{equation}
where 1 and 0 refer to the dimension over which the operation is applied.
This is similar to the MaxSim and sum operators used in ColBERT \cite{colbert-2020-khattab}, with the intuition that a relevant document should match as many query tokens as possible with the best-matching token.
The final relevance is a weighted sum over all heads:
\begin{equation*}
r(\bm{q}, \bm{d}) = \sum_{h=1}^{H}{P^{\text{head}}_h \cdot r_h(\bm{q}, \bm{d})},
\end{equation*}
where $P_h$ is a learnable weight that sums to one.
As explained in the next section, we empirically find only a few attention heads with non-random retrieval performance, and among them one particular head is significantly better than the others.
Given this observation, we introduce a low temperature $\tau$ to promote this sparsity $P^{\text{head}}_h = \frac{\exp(w_h / \tau)}{\sum_{h'}{\exp(w_h' / \tau)}}$, which always ends with \emph{a single head} with the great majority of the weight, which is denoted as \emph{retrieval head} $h^*$.
As a result, the learned head weights are practically a head selector, a fact that can also be exploited to make test-time retrieval more efficient.

\paragraph{End-to-end Retrieval with Attention}
To perform retrieval over a corpus, we first generate key vectors $K^{B+1,h^*}_{\bm{d}}$ of retrieval head for all document tokens offline and index them with FAISS library \cite{faiss-2021-johnson}.
For each query token, we issue its vector ($Q^{B+1,h^*}_{\bm{q}}$) to the index to retrieve top-$K'$ document tokens, which yields a filtered set of documents, each of which has at least one token retrieved by a query token.
We then fetch all tokens of filtered documents, compute relevance scores following \autoref{eq:avgmax}, and return top-$K$ documents with the highest scores $r_{h^*}(\bm{q}, \bm{d})$.
This is similar to the two-stage retrieval in ColBERT \cite{colbert-2020-khattab}, and we reuse their successful practice in index compression and search approximation to make test-time retrieval efficient, which we refer to \citet{colbertv2-2021-santhanam} for details.

\subsection{How Good is Attention As-is?}\label{sec:reatt_asis}
To examine this question, we use T5-large and test queries from the Natural Question dataset (NQ), retrieve 100 documents with BM25, compute relevance scores $r_h(\bm{q}, \bm{d})$ with half layers ($B=12$) as bi-encoder, and measure its correlation with the gold binary annotation.
We found that among $H=24$ heads, 4 heads have non-trivial correlations of 0.137, 0.097, 0.082, and 0.059.
We further perform end-to-end retrieval over Wikipedia using the best head, achieving top-10 retrieval accuracy of 43.5\%, inferior to 55.5\% of BM25.
This demonstrates that there are indeed heads that can relate queries with relevant documents, but they are not competitive.
We hypothesize that because self-attention is usually trained by comparing and relating tokens in a local context (512/1024 tokens) it cannot effectively identify relevant tokens in the enormous representation space of a corpus with millions of documents.
This discrepancy motivates us to compute retrieval attention between queries and potentially all documents (i.e., attention over the corpus), and \emph{adjust attention across documents to promote useful ones}.

\section{Learning Retrieval as Attention}\label{sec:learn_reatt}
We first approximate attention over the corpus at training time by sub-sampling a manageable number of documents for each query containing both potentially relevant and random documents (\autoref{sec:learn_reatt_aoc}).
Next, we introduce our end-to-end training objective that optimizes a standard QA loss while also adding supervision to promote attention over documents that are useful for the end task (\autoref{sec:learn_reatt_loss}).

\begin{figure}[tb]
\includegraphics[width=1.0\columnwidth, clip, keepaspectratio]{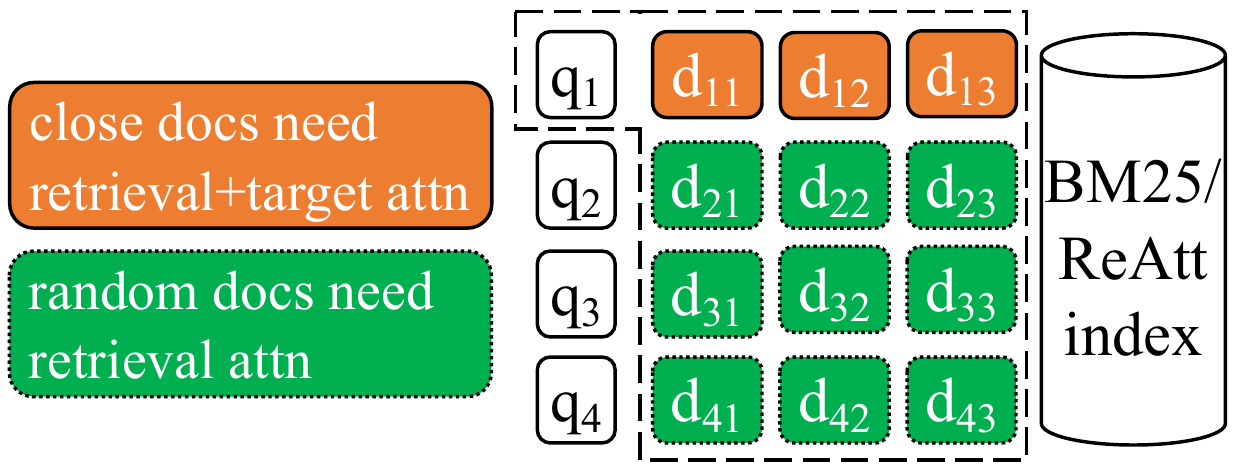}
\centering
\caption{Illustration of approximate attention over the corpus with $|\mathcal{Q}|$=4 queries in a batch and $K$=3 close documents per query. We use $\bm{q}_1$ as an example to illustrate the required computation, where \textcolor{Orange}{close documents} require both retrieval and target attention while \textcolor{Green}{random documents} only require retrieval attention.}
\label{fig:aoc}
\end{figure}

\subsection{Approximate Attention over the Corpus}\label{sec:learn_reatt_aoc}
Encoding the entire corpus and computing attention between the query and all documents is very expensive.
To make it practical, we propose to sub-sample a small set of documents for each query to approximate the whole corpus.
Inspired by negative sampling methods used in dense retriever training \cite{dpr-2020-karpukhin,ance-2021-xiong,colbert-2020-khattab}, we sub-sample both (1) documents close to queries that can be either relevant or hard negatives, and (2) random documents that are most likely to be easy negatives.
This allows the model to distinguish between relevant and hard negative documents, while simultaneously preventing it from losing its ability to distinguish easy negatives, which form the majority of the corpus.

\paragraph{Iterative Close Document Sub-sampling}
To sample documents close to a query $\mathcal{D}_{\bm{q}}^{\text{close}}$, we start from widely used lexical retriever BM25 \cite{bm25-2009-robertson} to retrieve $K=100$ documents, as shown by the orange blocks in \autoref{fig:aoc}.
We set $K$ to a relatively large number to better approximate the local region, inspired by \citet{fid-2021-izacard}'s findings that QA performance increases as more documents are used.

This fixed set of close documents can become outdated and no longer close to the query anymore as the retrieval attention gets better.
To provide dynamic close sub-samples, we re-index the corpus and retrieve a new set of $K$ documents using the current retrieval attention after each iteration.
It is similar in spirit to the hard negative mining methods used in \citet{dpr-2020-karpukhin,colbertnq-2020-khattab}, with a major difference that we do not manually or heuristically annotate documents but instead learn from the end loss with \adjustment, which will be explained in \autoref{sec:learn_reatt_loss}.

\paragraph{In-batch Random Document Sub-sampling}
We use close documents of other queries in the same batch as the random documents of the current query $\mathcal{D}_{\bm{q}}^{\text{random}}$ = $\cup_{\bm{q}' \in \mathcal{Q} \land \bm{q}' \neq \bm{q}}{\mathcal{D}_{\bm{q}'}^{\text{close}}}$ where $\mathcal{Q}$ contains all queries in a batch, as shown by the green blocks in \autoref{fig:aoc}, which has the advantage of reusing document representations across queries.
This is similar to the in-batch negatives used in DPR \cite{dpr-2020-karpukhin} with a major difference that we reuse a token representations ($K^{B+1,h}_{\bm{d}}, 1\leq h\leq H$) across queries instead of a single-vector document representation.

\subsection{Cross-document Adjustment with Decoder-to-Encoder Attention Distillation}\label{sec:learn_reatt_loss}
Given the sub-sampled $|\mathcal{Q}|\times K$ documents $\mathcal{D}_{\bm{q}} = \mathcal{D}_{\bm{q}}^{\text{close}} \cup \mathcal{D}_{\bm{q}}^{\text{random}}$ for each query $\bm{q}$, we compute the retrieval attention-based relevance scores $r(\bm{q},\bm{d})$ and \adjust them across multiple documents $\bm{d} \in \mathcal{D}_{\bm{q}}$ only relying on end task supervision.
Since retrieval is simply a means to achieve the downstream task, documents useful for downstream tasks should be promoted by retrieval.
Inspired by reader-to-retriever distillation \cite{fid-kd-2021-izacard,retread-2020-yang}, we measure document usefulness based on cross-attention between decoder and encoder, and minimize retrieval attention's discrepancy from it through distillation.
In contrast to \citet{fid-kd-2021-izacard} that learns two models iteratively and alternatively, we optimize QA and distillation loss in a single model simultaneously.

\paragraph{Minimizing KL-divergence Between Retrieval and Target Attention}
Specifically, we denote cross-attention before softmax of the first position/token of the last decoder layer as \emph{target attention} $C_{\bm{a}, \bm{q}, \mathcal{D}_{\bm{q}}} \in \mathbb{R}^{H \times |\mathcal{D}_{\bm{q}}| \times (|\bm{d}| + |\bm{q}|)}$ where $\bm{a}$ is the answer, $|\mathcal{D}_{\bm{q}}|$ is the number of sub-sampled documents to be fused by the decoder (\autoref{sec:reatt_definition}), and $|\bm{d}|$ is document length.\footnote{We also attempted other variations of target attention and found performances are similar, consistent with observations in \citet{fid-kd-2021-izacard}.}
To aggregate token-level target attention into document-level distribution $P^\text{tgt}(\bm{a}, \bm{q}, \mathcal{D}_{\bm{q}}) \in \mathbb{R}^{|\mathcal{D}_{\bm{q}}|}$, we first perform softmax over all tokens in all query-document pairs ($|\mathcal{D}_{\bm{q}}| \times (|\bm{d}|+|\bm{q}|)$), sum over tokens of each query-document pair ($|\bm{d}|+|\bm{q}|$), then average across multiple heads ($H$):
\begingroup\makeatletter\def\f@size{10}\check@mathfonts
\def\maketag@@@#1{\hbox{\m@th\large\normalfont#1}}
\begin{equation*}
P^\text{tgt}(\bm{a}, \bm{q}, \mathcal{D}_{\bm{q}}) = \text{avg}_{0}\Big(\text{sum}_{2}\big(\softmax_{1,2}(C_{\bm{a}, \bm{q}, \mathcal{D}_{\bm{q}}})\big)\Big).
\end{equation*}\endgroup
Given relevance scores obtained from retrieval attention, the final \adjustloss is the KL-divergence between relevance distribution $P^{\text{ret}}$ and target distribution $P^{\text{tgt}}$:
\begingroup\makeatletter\def\f@size{9}\check@mathfonts
\def\maketag@@@#1{\hbox{\m@th\large\normalfont#1}}
\begin{align}\label{eq:cal_loss}
\begin{split}
P^{\text{ret}}(\bm{q}, \mathcal{D}_{\bm{q}})&=\softmax\big(r(\bm{q}, \bm{d}_1),...,r(\bm{q}, \bm{d}_{|\mathcal{D}_{\bm{q}}|})\big). \\
\mathcal{L}_{\text{cross-doc}}&=\kl{\overline{P^\text{tgt}(\bm{a}, \bm{q}, \mathcal{D}_{\bm{q}})}}{P^{\text{ret}}(\bm{q}, \mathcal{D}_{\bm{q}})},
\end{split}
\end{align}\endgroup
where the overline indicates stop gradient back propagation to target distributions.
Our final loss combines QA loss and \adjustloss with $\alpha$ as combination weight.
\begin{equation}\label{eq:loss}
\mathcal{L} = \mathcal{L}_{\text{QA}} + \alpha \cdot \mathcal{L}_{\text{cross-doc}}.
\end{equation}

\paragraph{Zero Target Attention for Random Documents}
For a batch with $|\mathcal{Q}|$ queries, we need to compute retrieval attention and target attention between $|\mathcal{Q}|\times|\mathcal{Q}|\times K$ query-document pairs.
This is both computation- and memory-intensive when batch size is large, especially for target attention because it requires $L-B$ layers of joint encoding of query-document pairs in the cross-encoder.
To alleviate this, we make a simple and effective assumption that in-batch random documents are not relevant to the current query thus having zero target attention: $P^{\text{tgt}}(\bm{a}, \bm{q}, \mathcal{D}_{\bm{q}}^{\text{random}}) \in \mathbb{R}^{|\mathcal{D}_{\bm{q}}^{\text{random}}|} \gets 0$.
As a result, we only need to run cross-encoder and decoder for $K$ close documents of each query, as shown in \autoref{fig:aoc}.
In \autoref{app:eff_impl} we will introduce our efficient implementation to make it possible to run a large batch size over a limited number of GPUs.

\subsection{Domain Adaptation Methods}
One of the major benefits of a single end-to-end trainable model is that given a new corpus from a new domain, possibly without retrieval annotations, we can easily adapt it by end-to-end training. This section describes how we adapt \ours under different setups.

We consider adapting \ours with (1) QA supervision, (2) information retrieval (IR) supervision, or (3) unsupervised adaptation where we only have access to the document corpus.
Although our goal is to learn retrieval through downstream tasks instead of retrieval supervision, being able to consume retrieval annotations is helpful when retrieval supervision is indeed available.
To do so, we convert retrieval task with annotations in the form of query-document-relevance triples $\langle\bm{q}, \bm{d}, l\rangle$ into a generative task: given a query, the target is to generate \emph{its relevant document and the corresponding relevance} with the following format ``relevance: $l$. $\bm{d}$''.
If a query has multiple relevant documents, we follow \citet{fid-2021-izacard} to randomly sample one of them.
For unsupervised adaptation, with simplicity as our primary goal, we randomly choose one sentence from a document and mask one entity, which is considered as the ``query'', and have our model generate the masked entity as the ``answer'', similar to salient span masking (SSM) used in \citet{realm-2020-guu}.

\section{In-domain Experiments}
In this section, we examine if supervisedly training \ours end-to-end with \emph{only} QA supervision yields both competitive retrieval and QA performance.

\paragraph{Datasets, Baselines, and Metrics}
We train our model using the Natural Questions dataset (NQ).
We compare retrieval performance with lexical models BM25 \cite{bm25-2009-robertson}, passage-level dense retrievers DPR, ANCE, coCondenser, FiD-KD, YONO (with and without retrieval pretraining) \cite{dpr-2020-karpukhin,dpr-paq-2021-oguz,ance-2021-xiong,cocondenser-2022-gao,fid-kd-2021-izacard,yono-2021-lee}, and token/phrase-level dense retrievers DensePhrase, ColBERT, ColBERT-NQ \cite{densephraseret-2021-lee,colbert-2020-khattab,colbertnq-2020-khattab}.\footnote{\scriptsize ColBERT is trained on \msm, ColBERT-NQ is on NQ.}
Among them ColBERT-NQ, FiD-KD and YONO are the most fair-to-compare baselines because of either similar token-level retrieval granularity (ColBERT-NQ) or similar end-to-end training settings (FiD-KD and YONO).
We report top-k retrieval accuracy (R@k), the fraction of queries with at least one retrieved document containing answers.
We compare QA performance with ORQA, REALM, RAG, FiD, EMDR$^2$, YONO, UnitedQA, and R2-D2 \cite{orqa-2019-lee,realm-2020-guu,rag-2020-lewis,fid-2021-izacard,fid-kd-2021-izacard,emdr2-2021-sachan,yono-2021-lee,unitedqa-2021-cheng,r2d2-2021-fajcik} using exact match (EM), among which FiD, EMDR$^2$, and YONO are the most fair-to-compare baselines because they have similar model sizes and training settings.

\section{Implementation Details of \ours}\label{sec:impl_details}
\ours is based on T5-large with $B=12$ encoder layers as bi-encoder and temperatures $\tau=0.001$ to select the best retrieval head.
We retrieve $K=100$ close documents for each query, and use a batch size of $|\mathcal{Q}|=64$ queries to obtain in-batch random documents.
We use $\alpha=8$ to combine \adjustloss with QA loss.
We use AdamW with a learning rate of 5e-5, 10\% steps of warmup, and linear decay.
We first warmup cross-attention's ability to distinguish documents by only using the QA loss for 3K steps, then train with the combined losses (\autoref{eq:loss}) for 4 iterations, where the first iteration uses close documents returned by BM25, and the following 3 iterations use close documents returned by the previous \ours model (denoted as \ours$_\text{BM25}$).
Each iteration has 8K update steps and takes $\sim$ 1.5 days on a single node with 8 $\times$ A100 GPUs with 80GB memory.
Since DPR \cite{dpr-2020-karpukhin} achieves stronger performance than BM25, training with close documents returned by DPR can potentially reduce training time.
We experimented with training on close documents from DPR for a single iteration with 16K steps (denoted as \ours$_\text{DPR}$).
Since both approaches achieve similar performance (\autoref{tab:nq_ret} and \autoref{tab:nq_qa}) and \ours$_\text{DPR}$ is cheaper to train, we use it in other experimental settings.

At test-time, we save key vectors of all tokens in the corpus and use exact index from FAISS (i.e., \texttt{faiss.IndexFlatIP}) to perform inner-product search.
We retrieve $K'=2048$ document tokens for each query token and return top-100 documents with the highest aggregated scores (\autoref{eq:avgmax}) to generate answers.
We found compressing index with clustering and quantization proposed by \citet{colbertv2-2021-santhanam} can greatly reduce search latency and index size with a minor retrieval accuracy loss.

\subsection{Overall Results}
\begin{table}[tb]
\centering
\small
\begin{tabular}{l@{\middlecol}r@{\middlecol}r@{\middlecol}r@{\middlecol}r@{\middlecol}r}
\toprule
\textbf{Models} & \textbf{R@1} & \textbf{R@5} & \textbf{R@20} & \textbf{R@100} & \textbf{\#Params.} \\
\midrule
\multicolumn{6}{c}{\emph{supervised retrievers}} \\
BM25 & 23.9 & 45.9 & 63.8 & 78.9 & - \\
DPR & 45.9 & 68.1 & 80.0 & 85.9 & 220M \\
DPR$^{\text{new}}$ & 52.5 & 72.2 & 81.3 & 87.3 & 220M \\
DPR-PAQ & - & 74.2 & 84.0 & 89.2 & 220M \\
ANCE & - & - & 81.9 & 87.5 & 220M \\
coCondenser & - & 75.8 & 84.3 & 89.0 & 220M \\
DensePhrase & 51.1 & 69.9 & 78.7 & - & 330M \\
ColBERT & - & - & 79.1 & - & 110M \\
\hlrow ColBERT-NQ & 54.3 & 75.7 & 85.6 & 90.0 & 110M \\
\midrule
\multicolumn{6}{c}{\emph{semi/unsupervised retrievers}} \\
\hlrow FiD-KD & 49.4 & 73.8 & 84.3 & 89.3 & 220M \\
\hlrow YONO$_\text{w/o PT}$ & - & - & 72.3 & 82.2 & 165M \\
\hlrow YONO$_\text{w/ PT}$ & - & 75.3 & 85.2 & 90.2 & 165M \\
\ours$_\text{DPR}$ & 54.6 & 77.2 & \textbf{86.1} & \textbf{90.7} & 165M \\
\ours$_\text{BM25}$ & \textbf{55.8} & \textbf{77.4} & 86.0 & 90.4 & 165M \\
\bottomrule
\end{tabular}
\caption{Retrieval performance on NQ. PT is retrieval pretraining. Fair-to-compare baselines are highlighted with background color. Best performance is in bold.}
\label{tab:nq_ret}
\end{table}

\begin{table}[tb]
\centering
\small
\begin{tabular}{l@{\middlecol}r@{\middlecol}r}
\toprule
\textbf{Models} & \textbf{EM} & \textbf{\#Params.} \\
\midrule
ORQA \cite{orqa-2019-lee} & 33.3 & 330M \\
REALM \cite{realm-2020-guu} & 40.4 & 330M \\
RAG \cite{rag-2020-lewis} & 44.5 & 220M \\
\hlrow FiD \cite{fid-2021-izacard} & 51.4 & 990M \\
\hlrow FiD-KD \cite{fid-kd-2021-izacard} & 54.4 & 990M \\
\hlrow EMDR$^2$ \cite{emdr2-2021-sachan} & 52.5 & 440M \\
\hlrow YONO$_\text{w/o PT}$ \cite{yono-2021-lee} & 42.4 & 440M \\
\hlrow YONO$_\text{w/ PT}$ \cite{yono-2021-lee} & 53.2 & 440M \\
UnitedQA \cite{unitedqa-2021-cheng} & 54.7 & 1.870B\space\space \\
R2-D2 \cite{r2d2-2021-fajcik} & \textbf{55.9} & 1.290B\space\space \\
\ours$_\text{DPR}$ & 54.0 & 770M \\
\ours$_\text{BM25}$ & 54.7 & 770M \\
\bottomrule
\end{tabular}
\caption{QA performance on NQ. PT is retrieval pretraining. Fair-to-compare baselines are highlighted. Best performance is in bold.}
\label{tab:nq_qa}
\end{table}

We compare \ours with various retrievers and readers in \autoref{tab:nq_ret} and \autoref{tab:nq_qa}.
\ours achieves both slightly better retrieval performance than the strongest retriever baseline ColBERT-NQ \cite{colbertnq-2020-khattab} and comparable QA performance than the strong reader baseline FiD-KD \cite{fid-kd-2021-izacard} on NQ, demonstrating for the first time that fully end-to-end training using QA supervision can produce both competitive retrieval and QA performance.
Compared to another single-model architecture YONO \cite{yono-2021-lee}, \ours offers better performance without cumbersome pretraining to warm-up retrieval.

\subsection{Ablations}
We perform ablation experiments to understand the contribution of each component.
Due to resource limitations, all ablations are trained with 2K steps per iteration.
We use \ours trained with $B$=12 bi-encoder layers, $|\mathcal{Q}|$=16 batch size, and $\alpha$=8 cross-document loss weight as the baseline, remove one component or modify one hyperparameter at a time to investigate its effect.
As shown in \autoref{tab:abl}, we found: 1. Only using QA loss without \adjustment (\#2) improves retrieval performance over the original T5 (\#3), but \adjustment is necessary to achieve further improvement (\#1). 2. Iteratively retrieving close documents with the current model is helpful (\#5 vs \#1). 3. In-batch random documents are beneficial (\#4 vs \#1), and a larger batch size leads to larger improvements (\#8-11). 4. A larger weight on \adjustloss improves retrieval performance but hurts QA performance, with 4$\sim$8 achieving a good trade-off (\#12-15). 5. A small number of bi-encoder layers (\#6) significantly hurts retrieval while a large number of layers (\#7) significantly hurts QA, suggesting choosing equal numbers of layers in bi-encoder and cross-encoder.

\begin{table}[tb]
\centering
\small
\begin{tabular}{ll@{\smallcol}r@{\smallcol}r@{\smallcol}r@{\smallcol}r|@{\smallcol}r}
\toprule
\textbf{\#} & \textbf{Methods} & \textbf{R@1} & \textbf{R@5} & \textbf{R@20} & \textbf{R@100} & \textbf{EM} \\
\midrule 
& \multicolumn{6}{c}{\emph{\ours baseline with $B$=12, $|\mathcal{Q}|$=16, $\alpha$=8}} \\
1 & & 41.9 & 68.8 & 82.5 & 88.9 & 46.3 \\
\midrule
& \multicolumn{6}{c}{\emph{remove one component}} \\
2 & - cross-doc loss & 21.7 & 49.0 & 71.5 & 83.5 & 46.0 \\
3 & \quad - QA (=T5) & 13.2 & 33.7 & 53.6 & 67.7 & 3.0 \\
4 & - in-batch & 38.1 & 66.0 & 80.3 & 87.6 & 46.7 \\
5 & - iterative & 41.2 & 68.3 & 82.0 & 88.4 & 45.0 \\
\midrule
& \multicolumn{6}{c}{\emph{different \#layers in bi-encoder $B$}} \\
6 & $B$=6 & 19.1 & 42.1 & 62.4 & 78.1 & 40.3 \\
7 & $B$=18 & 38.2 & 63.8 & 79.3 & 87.4 & 35.2 \\
\midrule
& \multicolumn{6}{c}{\emph{different batch sizes $\mathcal{Q}$}} \\
8 & $|\mathcal{Q}|$=4 & 39.4 & 66.1 & 80.7 & 88.1 & 45.0 \\
9 & $|\mathcal{Q}|$=8 & 40.7 & 67.1 & 82.1 & 88.6 & 45.7 \\
10 & $|\mathcal{Q}|$=32 & 43.6 & 69.4 & 82.8 & 89.1 & 46.4 \\
11 & $|\mathcal{Q}|$=64 & 45.5 & 71.0 & 83.3 & 89.4 & 47.3 \\
\midrule
& \multicolumn{6}{c}{\emph{different cross-doc loss weights $\alpha$}} \\
12 & $\alpha$=1 & 37.4 & 65.4 & 80.9 & 88.0 & 47.3 \\
13 & $\alpha$=2 & 39.7 & 66.9 & 81.7 & 88.4 & 47.4 \\
14 & $\alpha$=4 & 40.9 & 68.0 & 82.1 & 88.8 & 46.9 \\
15 & $\alpha$=16 & 42.0 & 68.8 & 82.5 & 88.8 & 45.5 \\
\bottomrule
\end{tabular}
\caption{Ablations by removing one component or changing one hyperparameter from the \ours baseline.}
\label{tab:abl}
\end{table}

\section{Out-of-domain Generalization and Adaptation}
In this section, we examine both zero-shot retrieval performance on out-of-domain datasets and \ours's end-to-end adaptability in supervised (QA, IR) and unsupervised settings.

\subsection{Datasets, Baselines, and Metrics}
We choose 7 datasets from BEIR \cite{beir-2021-thakur}, a benchmark covering diverse domains and tasks.
On each dataset we compare \ours with different types of retrievers including BM25, DPR, and ColBERT.
We consider 2 QA datasets (BioASQ and FiQA \cite{bioasq-2015-tsatsaronis,fiqa-2018-maia}) and one IR dataset (\msm \cite{msmarco-2016-nguyen}) to evaluate supervised adaptation capability, and 4 other datasets (CQADupStack, TREC-COVID, SCIDOCS, SciFact \cite{cqa-2015-hoogeveen,treccovid-2020-voorhees,scidocs-2020-cohan,scifact-2020-wadden}) to evaluate unsupervised adaptation capability.
Detailed statistics are listed in \autoref{tab:datasets}.
We report nDCG@10 to measure retrieval performance and EM to measure QA performance.
We group all baselines into three categories and denote them with different colors in the following tables:
\begin{itemize}[leftmargin=*]
\item \textbf{Supervised adaptation models} are trained with downstream task supervision, including RAG trained on BioASQ, Contriever fine-tuned on FiQA, and docT5query, ANCE, ColBERT, and Contriever fine-tuned on \msm \cite{doct5query-2019-nogueira,ance-2021-xiong,colbert-2020-khattab,contriver-2021-izacard}.
\item \textbf{Unsupervised adaptation models} are trained on domain corpus in an unsupervised way such as contrastive learning or pseudo query generation, including SimCSE and TSDAE+GPL \cite{simcse-2021-gao,tsdae-2021-wang,gpl-2021-wang}.
\item \textbf{Pretraining models} are trained on corpora without direct exposure to the target domain, such as Contriever \cite{contriver-2021-izacard} trained with contrastive learning on Wikipedia and CCNet.
\end{itemize}
We highlight baselines in the same category as \ours in the following tables since comparison between them is relatively fair.
Details of adaptation of \ours can be found in \autoref{app:adapt_detail}.

\begin{table}[tb]
\centering
\small
\begin{tabular}{@{}l@{\middlecol}r@{\middlecol}r|l@{\middlecol}r@{}}
\toprule
\textbf{Tasks} & \multicolumn{2}{c|}{\textbf{QA}} & \multicolumn{2}{r}{\textbf{Retrieval}} \\
\textbf{Datasets} & \textbf{BioASQ} & \textbf{FiQA} & \multicolumn{2}{r}{\textbf{\msm}} \\
\midrule
\multicolumn{5}{c}{\emph{zero-shot performance}} \\
BM25 & 68.1 & 23.6 & & 22.8 \\
DPR & 14.1 & 11.2 & & 17.7 \\
\hlcell ColBERT-NQ & \hlcell 65.5 & \hlcell 23.8 & \hlcell & \hlcell \textbf{32.8} \\
\ours & \textbf{71.1} & \textbf{30.1} & & 32.3 \\
\midrule
\multicolumn{5}{c}{\emph{additional training}} \\
\pretrain{Contriever} & - & 32.9 & \hlcell \supervised{docT5query} & \hlcell 33.8 \\
\adaptation{SimCSE} & 58.1 & 31.4 & \hlcell \supervised{ANCE} & \hlcell 38.8 \\
\adaptation{TSDAE+GPL} & 61.6 & 34.4 & \hlcell \supervised{ColBERT} & \hlcell 40.1 \\
\hlcell \supervised{Contriever$_\text{w/ FT}$} & \hlcell - & \hlcell 38.1 & \hlcell \supervised{Contriever} & \hlcell \textbf{40.7} \\
\supervised{\ours} & \improve{5.8}\textbf{76.9} & \improve{8.5}\textbf{38.6} & \supervised{\ours} & \improve{7.6}39.9 \\
\bottomrule
\end{tabular}
\caption{nDCG@10 of zero-shot and supervised adaptation experiments on two QA and one IR datasets. We use colors to denote categories: \pretrain{pretraining}, \adaptation{unsupervised adaptation}, and \supervised{supervised adaptation}.
Baselines comparable to \ours are highlighted with blue background color. We also show the improvement of \ours over zero-shot performance in subscript.}
\label{tab:adapt_supervised}
\end{table}

\begin{table}[tb]
\centering
\small
\begin{tabular}{r@{\middlecol}l@{\middlecol}r@{\middlecol}r@{\middlecol}r|r@{\middlecol}r}
\toprule
\# & \textbf{Ablations} & \multicolumn{3}{l|}{{\footnotesize\textbf{nDCG@1 @5}}} & \multicolumn{2}{c}{\textbf{EM}} \\
\hline
\dlrow 1 & RAG & 14.6 & 13.0 &  & 1.3 & \\
2 & \squad+ reader & 14.6 & 13.0 & - & 27.5 & \better{26.2} \\
\hlrow 3 & \squad\squad+ qry enc (e2e) & 0.0 & 0.0 & \worse{-13.0} & 25.7 & \worse{-1.9} \\
4 & \squad+ doc/qry enc$^*$ & 29.4 & 27.1 & \better{14.1} & 5.0 & \better{3.7} \\
5 & \squad\squad+ reader (pipe) & 29.4 & 27.1 & - & 27.8 & \better{22.8} \\
6 & \squad\squad\squad+ qry enc & 23.3 & 23.2 & \worse{-4.0} & 26.2 & \worse{-1.6} \\
\hline
\dlrow 7 & T5 & 49.2 & 47.7 & & 0.0 & \\
\hlrow 8 & \squad+ e2e & 75.2 & 73.5 & \better{25.7} & 44.4 & \better{44.4} \\
\hline
\dlrow 9 & \ours & 72.8 & 70.1 & & 17.2 & \\
\hlrow 10 & \squad+ e2e & \textbf{77.4} & \textbf{75.4} & \better{5.3} & \textbf{47.2} & \better{30.0} \\
\hline
\end{tabular}
\caption{RAG and \ours on BioASQ. Each indent indicates fine-tuning one more component than its parent with performance difference colored with green/red. $^*$ denotes fine-tuning conducted sequentially instead of jointly with the current component.}
\label{tab:adapt_rag}
\end{table}

\begin{table}[tb]
\centering
\small
\begin{tabular}{l@{\middlecol}r@{\middlecol}r@{\middlecol}r@{\middlecol}r}
\toprule
\textbf{Methods} & \textbf{CQA.} & \textbf{TRECC.} & \textbf{SCIDOCS} & \textbf{SciFact} \\
\midrule
\multicolumn{5}{c}{\emph{zero-shot performance}} \\
BM25 & 29.9 & \textbf{65.6} & \textbf{15.8} & 66.5 \\
DPR & 15.3 & 33.2 & 7.7 & 31.8 \\
ANCE & 29.6 & 65.4 & 12.2 & 50.7 \\
\hlrow ColBERT-NQ & \textbf{33.9} & 48.9 & 15.6 & 65.3 \\
\ours & 33.3 & 62.6 & 14.8 & \textbf{71.0} \\
\midrule
\multicolumn{5}{c}{\emph{additional training}} \\
\pretrain{Contriever} & 34.5 & 59.6 & \textbf{16.5} & 67.7 \\
\hlrow \adaptation{SimCSE} & 29.0 & 68.3 & - & 55.0 \\
\hlrow \adaptation{TSDAE+GPL} & 35.1 & 74.6 & - & 68.9 \\
\adaptation{\ours} & \improve{3.3}\textbf{36.6} & \improve{13.4}\textbf{76.0} & \improve{1.0}15.8 & \improve{0.2}\textbf{71.2} \\
\bottomrule
\end{tabular}
\caption{nDCG@10 of zero-shot and unsupervised adaptation on four datasets. Format is similar to \autoref{tab:adapt_supervised}}
\label{tab:adapt_unsupervised}
\end{table}

\subsection{Experimental Results}
Results of supervised and unsupervised adaptation are listed in \autoref{tab:adapt_supervised}, \autoref{tab:adapt_rag}, and \autoref{tab:adapt_unsupervised} respectively.

\paragraph{Zero-shot Generalization Ability}
As shown in \autoref{tab:adapt_supervised} and \autoref{tab:adapt_unsupervised}, the zero-shot performance of \ours is significantly better than other zero-shot baselines on two QA datasets and one fact checking dataset (+3.0/+6.5/+4.5 on BioASQ/FiQA/SciFact than the second best), and overall comparable on the rest of datasets (-0.5/-0.6/-3.0/-1.0 on \msm/CQA./TRECC./SCIDOCS than the best which is usually BM25), demonstrating that our end-to-end training with QA loss on NQ produces a robust retriever.
We conjecture that the superior performance on QA datasets can be attributed to our end-to-end training using QA loss which learns retrieval that better aligns with the end task than training with retrieval annotations.

\paragraph{Retrieval Adaptation with QA Supervision}
As shown in the left-hand side of \autoref{tab:adapt_supervised}, end-to-end adaptation with QA supervision significantly improves \ours's retrieval performance by 5.8/8.5 on BioASQ/FiQA, achieving similar performance as Contriever fine-tuned on FiQA, and better performance than other unsupervised methods, confirming the end-to-end adaptability of our methods.

\paragraph{End-to-end QA Adaptation}
We perform end-to-end adaptation on BioASQ and compare with RAG as a baseline, which combines DPR as retriever and BART as reader, and DPR has a query and document encoder.
Since updating document encoder requires corpus re-indexing, it is fixed during fine-tuning.
We found end-to-end fine-tuning fails on RAG.
To understand why, we conduct a rigorous experiment that breaks down each component of RAG to find the failure point in \autoref{tab:adapt_rag}.

Starting from the initial model trained on NQ (\#1), we first fine-tune the reader while fixing the query encoder (\#2), and as expected QA performance improves.
However fine-tuning both query encoder and reader (end-to-end \#3) makes the retriever collapse with zero relevant documents returned, indicating end-to-end fine-tuning does not work for RAG on new domains.
In order to improve both retrieval and QA, we need to fine-tune RAG in a pipeline manner: first fine-tune the retriever (both query and doc encoder) similarly to DPR using retrieval annotations (\#4), then fine-tune the reader (\#5).
With the DPR-like fine-tuned retriever, end-to-end fine-tuning of query encoder and reader still fails (\#6), although the retriever does not completely collapse.

End-to-end fine-tuning of \ours improves retrieval and QA simultaneously.
Fine-tuning starting from \ours trained on NQ is better than starting from T5, indicating the capability learned in NQ could be transferred to BioASQ.
Comparing RAG and \ours, we identify several keys that enable end-to-end adaptation.
(1) \ours relying on token-level attention has a strong initial performance, (2) \adjustment over both close and random documents in \ours provides a better gradient estimation than only using retrieved documents in RAG, (3) distillation-based loss in \ours might be more effective than multiplying the retrieval probability into the final generation probability.

\paragraph{Leveraging Retrieval Annotations}
As shown on the right-hand side of \autoref{tab:adapt_supervised}, \ours is able to consume retrieval supervision in a generative format and achieve competitive performance as other supervised dense retrievers.

\paragraph{Unsupervised Adaptation with SSM}
As shown in \autoref{tab:adapt_unsupervised}, adaptation by simply masking salient entities from sentences as input and generating masked entities using \ours improves the retrieval performance on 4 datasets, some by a large margin, achieving comparable or superior performance than strong retrieval adaptation methods such as TSDAE+GPL that relies on query generation.
This indicates that our end-to-end trainable model also works well in unsupervised settings without involving too many engineering heuristics.

\section{Related Work}\label{sec:related_work}
Retrieval-augmented question answering utilizes evidence retrieved from an external knowledge source to facilitate question answering.
There have been several attempts to learn retrievers and readers jointly.
ORQA, REALM, RAG, EMDR$^2$, YONO, and Atlas \cite{orqa-2019-lee,realm-2020-guu,emdr2-2021-sachan,yono-2021-lee,atlas-2022-izacard} first warm-up retrievers using unsupervised pretraining methods such as inverse cloze task (ICT), salient span masking (SSM), and large-scale contrastive learning, or initialize from supervised retrievers, then fine-tune both retriever and reader on downstream tasks.
They either use fixed index \cite{orqa-2019-lee,rag-2020-lewis} or asynchronously update the index during training \cite{realm-2020-guu,emdr2-2021-sachan,yono-2021-lee,atlas-2022-izacard}.
Recently, retrieval-augmented models are scaled up to very large corpora such as the web \cite{oyster-2021-piktus,retro-2021-retro}, making them capable of handling information out of the scope of Wikipedia.
Atlas \cite{atlas-2022-izacard} scales up retrieval-augmented models with T5-11B as the reader and Contriever \cite{contriver-2021-izacard} as the retriever and achieves strong few-shot performance on multiple benchmarks.
Detailed comparisons of these models can be found in \autoref{tab:retaug_related}.
More related works on dense retrieval, unsupervised retrieval learning, and retrieval augmentation for language modeling can be found in \autoref{app:related_work}.

\section{Conclusion}
We propose retrieval as attention (\ours), a single Transformer model that can be learned in an end-to-end fashion only using end task loss.
We demonstrated on NQ dataset that \ours can achieve both competitive retrieval and QA performance.
We further show that \ours is easy to adapt to other domains in both supervised and unsupervised settings, achieving both boosted retrieval and end task performance.
Future directions include better end-to-end training objectives and efficient training and inference.

\section*{Limitations}
\ours is based on token-level representations, and belongs to the same category as token-level dense retrievers such as ColBERT \cite{colbert-2020-khattab}.
Comparing to passage-level dense retrievers such as DPR \cite{dpr-2020-karpukhin}, token-level retrievers usually offer better performance (shown in \autoref{tab:nq_ret}, \autoref{tab:adapt_supervised}, and \autoref{tab:adapt_unsupervised}) but require more space to store the index and longer query time.
Our methods have the same limitation.
We found ColBERT's practice in index compression and approximate search \cite{colbert-2020-khattab,colbertv2-2021-santhanam,plaid-2022-santhanam} also works for our model, making this issue less of a concern.

\section*{Acknowledgments}
This work was supported by a gift from Bosch Research.
We would like to thank Chunting Zhou, Uri Alon, Omar Khattab, Patrick Lewis, and Jane Dwivedi-Yu for their insightful feedback and help with experiments.

\bibliography{anthology,custom}
\bibliographystyle{acl_natbib}

\appendix
\begin{table*}[tb]
\centering
\small
\begin{tabular}{l|l@{\tinycol}l@{\tinycol}l|l@{\tinycol}l@{\tinycol}l|l}
\toprule
\multirow{2}{*}{\textbf{Model}} & \multicolumn{3}{c|}{\textbf{Architecture}} & \multicolumn{3}{c|}{\textbf{Retriever training}} & \multirow{2}{*}{\textbf{Granu.}} \\
 & \textbf{Retriever} & \textbf{Reader} & \textbf{Single} & \textbf{Init.} & \textbf{Warm-up} & \multicolumn{1}{c|}{\textbf{End-to-end loss}} & \\
\midrule
ORQA \cite{orqa-2019-lee} & BERT & BERT & \xmark & BERT & ICT & Prob. marginalization & Passage \\
REALM \cite{realm-2020-guu} & BERT & BERT & \xmark & BERT & ICT, SSM & Prob. marginalization & Passage \\
RAG \cite{rag-2020-lewis} & BERT & BART & \xmark & DPR & - & Prob. marginalization & Passage \\
EMDR$^2$ \cite{emdr2-2021-sachan} & BERT & T5 & \xmark & BERT & ICT, SSM & Expectation maximization & Passage \\
YONO \cite{yono-2021-lee} & T5 & T5 & \cmark & T5 & SSM & Attention distillation & Passage \\
Atlas \cite{atlas-2022-izacard} & BERT & T5 & \xmark & Contriever & MLM & Perplexity distillation & Passage \\
\midrule
\ours & T5 & T5 & \cmark & T5 & - & Attention distillation & Token \\
\bottomrule
\end{tabular}
\caption{Detailed comparison between end-to-end retriever-reader models. ICT is inverse cloze task, SSM is salient span masking, and MLM is masked language modeling. Granu. is retrieval granularity.}
\label{tab:retaug_related}
\end{table*}

\begin{table*}[tb]
\centering
\small
\begin{tabular}{l|l|l|rr|rr}
\toprule
\multirow{2}{*}{\textbf{Dataset}} & \multirow{2}{*}{\textbf{Domain}} & \multirow{2}{*}{\textbf{Task}} & \multicolumn{2}{c|}{\textbf{Train}} & \multicolumn{2}{c}{\textbf{Test}} \\
 & & & \textbf{\#Queries} & \textbf{\#Annotations} & \textbf{\#Queries} & \textbf{|Corpus|} \\
\hline
\multicolumn{7}{c}{\emph{In-domain}} \\
NQ & Wiki & Question answering & 79K & 133K & 3,610 & 21.015M \\
\midrule
\multicolumn{7}{c}{\emph{Out-of-domain supervised adaptation}} \\
BioASQ & Biomed & Question answering & 3K & 32K & 500 & 1.000M \\
FiQA & Finance & Question answering & 6K & 14K & 648 & 58K \\
\msm & Misc & Information retrieval & 503K & 533K & 6,980 & 8.842M \\
\midrule
\multicolumn{7}{c}{\emph{Out-of-domain unsupervised adaptation}} \\
CQADupStack & StackExchange & Duplicate question retrieval & - & - & 13,145 & 457K \\
TREC-COVID & Biomed & Information retrieval & - & - & 50 & 171K \\
SCIDOCS & Science & Citation prediction & - & - & 1,000 & 26K \\
SciFact & Science & Fact checking & 1K & 1K & 300 & 5K \\
\bottomrule
\end{tabular}
\caption{Statistics of 8 datasets categorized by experimental settings, including the number of training/test queries, retrieval annotations (query-document pairs), and documents in the corpus.}
\label{tab:datasets}
\end{table*}

\newpage

\section{Efficient Implementation}\label{app:eff_impl}
Under typical optimization setups where the loss is point-wise with respect to each training data, like training classifiers or readers, scaling batch size can be easily achieved with gradient accumulation. However, due to the use of in-batch negatives, our systems, like others~\cite{dpr-2020-karpukhin, rocketqa-2021-qu}, require having all examples in a batch to reside in GPUs simultaneously when trained directly. Larger batches therefore need proportionally more GPU memory.

In order to accommodate large batches with our limited memory hardware, we adopt the gradient cache approach~\cite{gao-etal-2021-scaling} decouple instances in the same batch. In particular, we run an extra forward pass over the large batch in inference mode and record (1) representations for all query and document tokens ($Q^{B+1,h}_{\bm{q}}$ and ${K^{B+1,h}_{\bm{d}}}$) and (2) decoder-encoder target attention values ($C_{\bm{a}, \bm{q}, \mathcal{D}_{\bm{q}}}$). Note that we \emph{do not} store model internal activation nor perform gradient computation with respect to model parameters in this step. With (1) we can compute the retrieval attention, and with (2) we can compute \adjustloss (\autoref{eq:cal_loss}). We then compute and cache gradient vectors of all query and document vectors with respect to \autoref{eq:cal_loss}. We finally optimize the model with a sufficiently small batch size to fit in GPU memory and use cached gradient in the backward pass of the \autoref{eq:cal_loss}.

\section{Details of Adaptation Experiments}\label{app:adapt_detail}
For supervised adaptation, we train on BioASQ, FiQA, and \msm separately using all training queries.
For CQADupStack, we merge the document corpora of 12 sub-domains into a single corpus to sample masked sentences for salient span masking training.
For each of the 4 unsupervised domain adaptation datasets (CQADupStack, TREC-COVID, SCIDOCS, SciFact), we sample 20$\sim$100K sentences and mask one entity, which is approximately proportional to the size of the corpus with a larger sampling rate for small corpora.
We reuse the same hyperparameters as NQ (\autoref{sec:impl_details}), except that we train each model for a single iteration using close documents from BM25 with 4K update steps and a batch size of 16.
Since \msm has a large number of annotations, we train for 12K update steps.

\section{Related Work}\label{app:related_work}
\paragraph{Dense Retrieval Models}
Dense retrieval models can be categorized into two groups, passage-level retrievers \cite{dpr-2020-karpukhin,dpr-paq-2021-oguz,ance-2021-xiong,cocondenser-2022-gao} and token/phrase-level retrievers \cite{colbert-2020-khattab,colbertnq-2020-khattab,coil-2021-gao,densephraseret-2021-lee}.
Passage-level retrievers encode queries and documents into a single vector, while token/phrase-level retrievers directly use token/phrase representations, resulting in multi-vector representations.
Passage-level retrievers are usually more efficient but less expressive than token-level retrievers.

\paragraph{Unsupervised Retrieval Learning}
Unsupervised retrieval learning methods can be categorized into two types: pretraining-based \cite{orqa-2019-lee,simcse-2021-gao,tsdae-2021-wang,condenser-2021-gao,contriver-2021-izacard} and question generation-based \cite{qgen-2021-ma,gpl-2021-wang}.
SimCSE \cite{simcse-2021-gao} obtains representations of the same input by passing through the model twice with different dropout masks and minimizes their distance.
Contriever \cite{contriver-2021-izacard} is trained by large-scale contrastive learning with random cropping of text spans sampled from Wikipedia and CCNet.
GPL \cite{gpl-2021-wang} leverages query generators to obtain pseudo queries, and collect positive and negative documents by pseudo labeling using a cross-encoder.

\paragraph{Retrieval Augmentation for Language Modeling}
Retrieval from external datastore to improve language modeling perplexity has been explored by many works, where additional tokens are retrieved during generation based on contextual representations \cite{knnlm-2020-khandelwal,retro-2021-retro,memtrans-2022-wu,trime-2022-zhong}.
They differ in whether retrieval is fixed or learnable, retrieval frequency, and contextual representations used to perform nearest neighbors search.

\end{document}